\documentclass[lettersize,journal]{IEEEtran}
\usepackage{amsmath,amsfonts}
\usepackage{algorithmic}
\usepackage{algorithm}
\usepackage{array}
\usepackage[caption=false,font=normalsize,labelfont=sf,textfont=sf]{subfig}
\usepackage{textcomp}
\usepackage{stfloats}
\usepackage{url}
\usepackage{verbatim}
\usepackage{graphicx}
\usepackage{xcolor}
\usepackage{cite}
\usepackage{caption} 
\usepackage{amssymb}
\usepackage{authblk}

\usepackage[hidelinks]{hyperref}
\makeatletter
\def\blfootnote{\xdef\@thefnmark{}\@footnotetext}
\makeatother

\begin{document}

\title{Tiny-HR: Towards an interpretable machine learning pipeline for heart rate estimation on edge devices
}

\author[1,*]{Preetam Anbukarasu}
\author[2]{Shailesh Nanisetty}
\author[1]{Ganesh Tata}
\author[1]{Nilanjan Ray}
\affil[1]{Department of Computing Science, University of Alberta, Edmonton, Alberta, Canada}
\affil[2]{Department Of Industrial and Systems Engineering, IIT Kharagpur, West Bengal, India}
\affil[*]{Corresponding Author: Preetam Anbukarasu, anbukara@ualberta.ca}


\markboth{IEEE Transactions On Consumer Electronics,~Vol.~\#, No.~\#, August 2022 (Preprint)}%
{Shell \MakeLowercase{\textit{et al.}}: A Sample Article Using IEEEtran.cls for IEEE Journals}

\IEEEoverridecommandlockouts
\IEEEpubid{\makebox[\columnwidth]{978-1-5386-5541-2/18/\$31.00~\copyright~2022 IEEE \hfill}
\hspace{\columnsep}\makebox[\columnwidth]{ }}

\maketitle

\IEEEpubidadjcol

\begin{abstract}
The focus of this paper is a proof of concept, machine learning (ML) pipeline that extracts heart rate from pressure sensor data acquired on low-power edge devices. The ML pipeline consists an upsampler neural network, a signal quality classifier, and a 1D-convolutional neural network optimized for efficient and accurate heart rate estimation. The models were designed so the pipeline was less than 40 kB. Further, a hybrid pipeline consisting of the upsampler and classifier, followed by a peak detection algorithm was developed. The pipelines were deployed on ESP32 edge device and benchmarked against signal processing to determine the energy usage, and inference times. The results indicate that the proposed ML and hybrid pipeline reduces energy and time per inference by 82\% and 28\% compared to traditional algorithms. The main trade-off for  ML pipeline was accuracy, with a mean absolute error (MAE) of 3.28, compared to 2.39 and 1.17 for the hybrid and signal processing pipelines. The ML models thus show promise for deployment in energy and computationally constrained devices. Further, the lower sampling rate and computational requirements for the ML pipeline could enable custom hardware solutions to reduce the cost and energy needs of wearable devices.

\end{abstract}


\section{Introduction}

Embedded biometric and wearable devices are some of the fastest growing consumer electronic devices because of their versatility and ease of use for remote health care \cite{bayoumy2021smart}, authentication \cite{Karimian2019} and tracking. With over 250 million devices deployed, they are a rich source of health data that usually do not get utilized to their full potential \cite{martinez2020quality}, due to limitations imposed by processing power, bandwidth, compliance and energy constraints of edge devices. In addition, the limited footprint and cost constraints make it impractical to integrate more powerful processing units and larger batteries in these devices. The vast majority of the memory, computation and consequently the energy consumption in these devices arise from the processing units and data acquisition system. Prior studies have shown that the sampling rate is a major factor in the baseload power consumption (up to 30\%) in biometric sensors \cite{tobola2015sampling}. A simple solution to reduce power consumption in wearable devices is to decrease the sampling rate and to put the data acquisition and processing unit in sleep mode between samples. However, this approach is limited by the initialization time and initialization energy of the data acquisition systems ($\sim$10’s of µS and 1 mA) and the Nyquist frequency of the features obtained from the signal, since the sampling rate has to be high enough for reliable signal reconstruction using traditional signal processing techniques. The generally accepted minimum sampling rate for heart rate estimation from wearable sensors is 10-20 Hz \cite{beres2019minimal}, with 12 Hz being at the threshold below which the accuracy of the heart rate estimation drops off markedly \cite{wolling2018fewer}. Studies have shown that low sampling rates can still produce reliable HR estimates if the signal-to-noise ratio is low and a suitable non-linear interpolation is used to reconstruct the signal \cite{zaunseder2022signal}. However, it is still typical for the sampling rate to be 2 to 8 times higher than Nyquist frequency to ensure sufficient signal quality. Due to these high sampling requirements, the energy saved by putting the ADC and processing unit in sleep mode between samples is diminished substantially. The signal processing pipeline, on the other hand, is responsible for 30-50\% of the mean energy and memory usage in the processing unit, with the inference time and energy used by the various signal conditioning and feature extraction algorithms scaling at least linearly with sampling rate \cite{tobola2015sampling}. As a result, the selection criteria for embedded hardware and microprocessor units are constrained by the sampling rate and the complexity of the signal processing pipeline required to extract relevant features for inference. These challenges increase the cost and complexity of the hardware while reducing the inference efficiency of wearable edge devices.

In recent years, multiple studies have demonstrated that machine learning approaches, especially deep or convolutional neural networks (DNN and CNN) \cite{acharya2017deep,8682194,bursa2017use,wen2019time} and signal classifiers \cite{lee2019fast,soltane2004artificial,liu2020classification,li2012dynamic,roy2020photoplethysmogram}, show great promise in augmenting the capabilities of traditional signal processing approaches for biomedical embedded devices \cite{Maqsood2021, schmidt2019peak,coutts2020deep}. Several research studies have shown the potential of deep learning models for applications like heart rate detection from photoplethysmogram \cite{Karimian2019,Reiss2019,Risso2021,Staffini2022,chang2021deepheart}, heart anomaly detection \cite{9669005,vsabic2021healthcare,8852242}, and motion artifact tolerant heart rate extraction and signal reconstruction \cite{li2012dynamic,shin2022deep,liu2020classification,roy2018improving,chowdhury2019robust,svantesson2021virtual}. Further, these DNNs are being optimized by approaches such as, sparsification, pruning, and quantization, which have reduced the size and complexity of these models without compromising their accuracy and usefulness \cite{heim2021measuring}. TinyML tools such as TensorFlow Lite and Cube AI have enabled the conversion of trained deep learning models to enable deployment in embedded devices with limited program memory, RAM and processing power \cite{ray2021review,han2022tinyml,banbury2020benchmarking,osman2022tinyml}.
 
Despite these advances, TinyML is still in its infancy and compact DNN/CNN models have not been extensively used in consumer devices and embedded systems, especially for critical applications where a false inference or misclassification can have major ramifications and liabilities. There are two main reasons for the lack of real-world applications, 1. Most DNNs are black boxes with very limited interpretability, 2. These DNNs are prohibitively large and energy-hungry to be deployed on embedded devices, especially when dealing with high sampling rate data.

This paper focuses on a lightweight ML pipeline that addresses the issues of interpretability and deployment on edge devices, by using three distinct models: 1. An upsampler neural network (U-NN) that takes low sampling rate sensor data as input and reconstructs it into a filtered, higher sampling rate output, 2. A classifier that is trained to detect abnormal, low-quality signals to process independently of the good quality signals and 3. A heart rate regression convolutional neural network (HR-NN) that estimates heart rate from the good quality signals obtained from the classifier. All parts of the pipeline are optimized for deployment in low-powered edge devices and the output of each ML model that forms the pipeline is designed to be probable to improve interpretability. We compared the ML pipeline with a fully interpretable hybrid pipeline (U-NN + Classifier + Traditional peak detection based HR estimator) and a traditional signal processing pipeline (Signal conditioning + Noise reduction + Peak Detection). We demonstrate that the proposed three-part ML pipeline performs favourably in terms of inference time while being slightly inferior in terms of accuracy. The results further indicate that the  ML pipeline deployed on ESP32, while having a larger model size, is more energy efficient than signal processing. The traditional signal processing pipeline resulted in a relatively higher accuracy at the expense of much longer inference times. The hybrid pipeline showed comparable accuracy with a lower energy consumption than the signal processing pipeline.

\section{Background}

\subsection {Upsampling Neural Networks for biometric signal}

Upsampling is one of the simplest ways to restore or reconstruct a signal to its original version from a low-resolution input. Upsampling neural networks (U-NN) have been used extensively in image super-resolution \cite{7115171,9044873}, medical imaging in applications such as FastMRI using compressed sensing \cite{9703109,9792767} and EEG signal reconstruction \cite{9747439}. The main advantage of the U-NN versus interpolation is the relative fidelity and the recovery of finer features of the signal making them more representative of the ground truth. In this study, we use a U-NN to reconstruct a 12 Hz output signal from a 6 HZ input, which can reduce the true sampling rate, energy consumption and duty cycle of the data acquisition system. By putting the ADC to sleep for the time between samples, we can reduce the ADC energy consumption by 50\% without compromising on signal quality. In addition to reconstruction, the U-NN effectively performs the function of signal conditioning and filtering algorithms as well. We demonstrate that the U-NN can reconstruct usable signals from input signals obtained at sub-optimal sampling space below the Nyquist rate.

\subsection{Interpretable Machine Learning and Signal Classifier}

Most DNNs are black boxes with limited interpretability, especially if their architecture is complex with multiple hidden layers and nodes. Interpretability is of paramount importance for mission-critical applications as in medical devices that produce inferences that would be used for diagnosis. Many previous works have proposed potential approaches to improve the interpretability and robustness of deep learning models \cite{funer2022accuracy,salahuddin2022transparency}. For our application, the interpretability of the pipeline arises from a decision-making classifier that is trained with expert labelled datasets to identify abnormal or poor-quality signals that are not suitable for further processing. By creating a go/no-go gate using the classifier, we avoid the possibility of inaccurate inferences and wasting real-time processing resources in the embedded device on poor-quality data. From an interpretability standpoint, the classifier gate provides insights into the location of the pipeline at which the model might fail when deployed in an environment where data sets that were not encountered during training. For example, the model can encounter situations where the input data itself contains a lot of noise and motion artifacts, in which case the signal may have to be conditioned before processing or even dropped altogether. Alternatively, an issue with the upsampling step can be flagged and investigated by a human expert at the classifier gate. We have discussed different scenarios wherein the output characteristics of the upsampler, classifier and heart rate regressor are exposed to abnormal datasets and noisy input signals and the potential changes in the estimation accuracy and reliability of the proposed models. We further discuss the potential limitations and future directions of studies to address these limitations.

\subsection{Machine Learning for Heart Rate Estimation}

\begin{figure*} [!h]
\centering
\includegraphics[width=\linewidth]{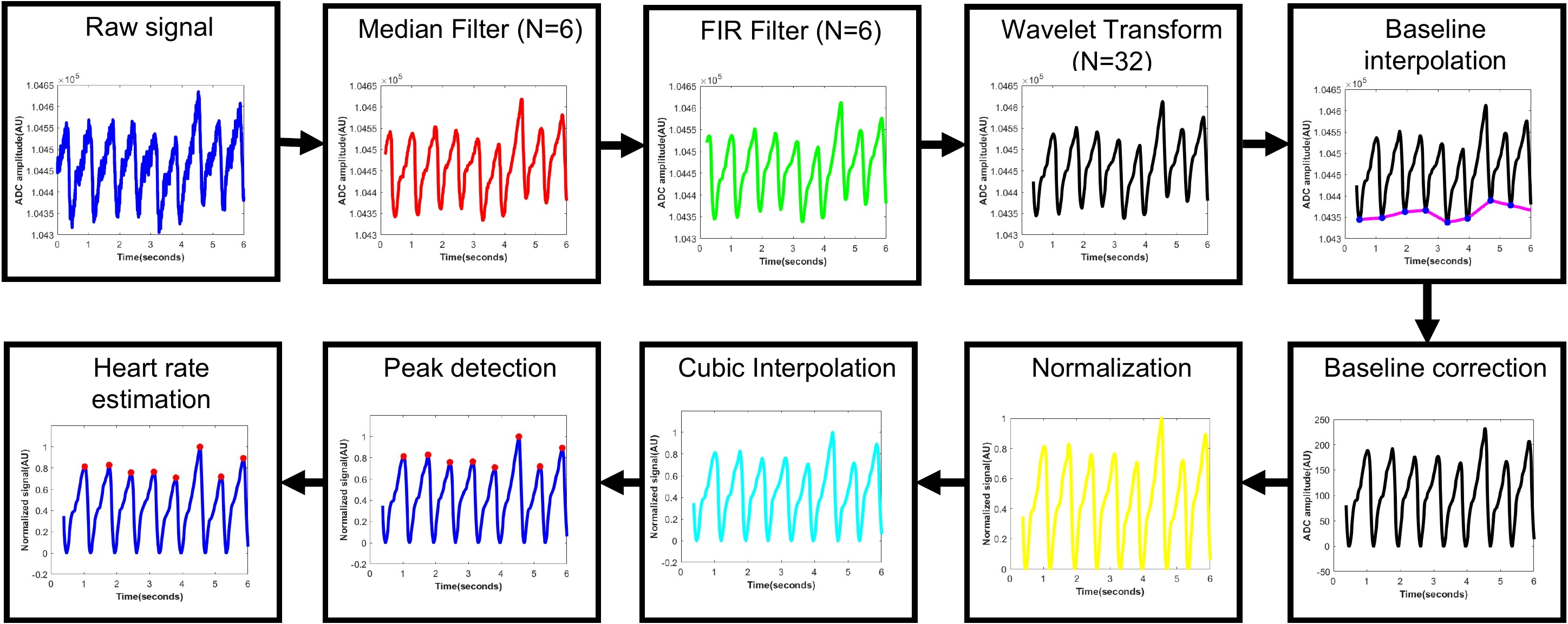}
\caption{Traditional signal processing pipeline used for comparison in the paper. The pipeline consists of multiple stages of signal conditioning filters, smoothing functions and peak detection algorithms to estimate heart rate.}
\label{Fig_1:tradition_pipeline}
\end{figure*}

Heart rate (HR) monitoring in wearable devices is typically carried out using wrist photoplethysmogram (PPG) signals. PPG data quality is affected by a range of factors including motion artifacts, skin tone variations and hydration levels, etc \cite{puranen2020effect,dasari2021evaluation}. In the current study, a tactile pressure sensor is used to obtain the pulse wave data from which the heart rate is estimated. The tactile pressure sensor affords a more robust signal with a higher signal-to-noise ratio, which is not as badly affected by environmental factors. Typically, the heart rate from wearable sensors is estimated using an extensive signal processing pipeline consisting of moving average filters, baseline correction algorithms, thresholding algorithms and peak detection algorithms. Fig.\ref{Fig_1:tradition_pipeline} is an example of a signal processing pipeline used to extract biometrical parameters from the raw signal. These algorithms are well studied and are currently used in embedded devices, even though they consume considerable inference time and energy for HR estimation. In this study, we have explored an HR regression neural network that takes conditioned and upsampled 12 Hz data as input and provides windowed mean HR values as output. We have further compared inference times and energy of the HR-NN versus the traditional pipeline. The ML pipeline was found to be more energy efficient while being slightly more inaccurate than traditional and hybrid pipelines.

\section{Method}
\subsection{Data Acquisition and Processing}
The raw pulse data used for the study was obtained from a custom-made, thin film sensor applied over the radial artery on the wrist. The data were acquired at 126 Hz using an ESP32 edge device equipped with an external 16-bit ADC and a programmable gain amplifier. The sensitivity and dynamic range of the sensors were optimized to ensure the reliable acquisition of the pressure pulse data. The signals were segmented into 6-second frames, then baseline corrected and normalized to ensure standardized input for the signal processing algorithms and the neural networks. The datasets were downsampled to an effective sampling rate of 6 Hz and 12 Hz to use as inputs and ground truth for the upsampling neural networks, while the 12 Hz datasets were used as input for the signal processing algorithms. The heart rate ground truth was determined by measuring the peak to peak time and by averaging the values over the 6-second interval for each dataset. The datasets were labelled as either good quality signals or poor quality signals, as determined by the signal quality, presence of motion artifacts, relative amplitude and peak prominence. The labelled datasets were used to train a signal quality classifier to prevent HR regressor inferences from unsuitable low-quality signals. In total, 5687 labelled signals were obtained from 5 healthy volunteers for the study. The datasets were acquired with the volunteers in seated position at rest or after moderate exercise.

The datasets were passed through a traditional signal processing pipeline consisting of signal filters, noise reduction and signal conditioning algorithms, to determine the HR ground truth. Fig. \ref{Fig_1:tradition_pipeline}. shows the major part of the pipeline used in this study to generate the HR values. Specifically, a combination of 6-order median and FIR filters were used to remove high-frequency noise and fluctuations in the samples. Further, baseline correction was performed to remove the oscillations and breathing artifacts present in the signal. Normalization was carried out to standardize the datasets and to make them comparable across samples. The normalized curves were fed to a peak detection algorithm to determine the mean heart rate for each data frame. The ground truth values were randomly sampled for different datasets and independently validated by two experts (Biomedical engineers) to be within a mean absolute error (MAE) of 0.24, which is comparable to the best signal processing algorithms available for HR prediction.

\subsection{Signal Processing Pipeline}
For edge deployment, we optimized the pipeline shown in Fig. \ref{Fig_1:tradition_pipeline} for lower model size, inference times and fewer overall steps to enable a fair comparison with the machine learning and hybrid pipelines discussed in the paper. The lightweight signal processing pipeline consists of 6-order median and FIR filters, baseline correction, followed by normalization. The normalized signals were interpolated by a factor of 10 to enable more accurate HR estimates. The interpolated curves were smoothed by a 30-order moving average filter before being fed into a peak detection algorithm to determine the mean heart rate for each dataset. The peak detection algorithm was developed based on previous studies \cite{baek2017reliability,beres2021minimal} on energy-efficient heart rate detection that uses a 3-point non-linear interpolator after the approximate peak location is determined to reduce inference time and computational requirements.

\subsection{Machine Learning Pipeline}

\begin{figure*} [!h]
\centering
\includegraphics[width=\linewidth]{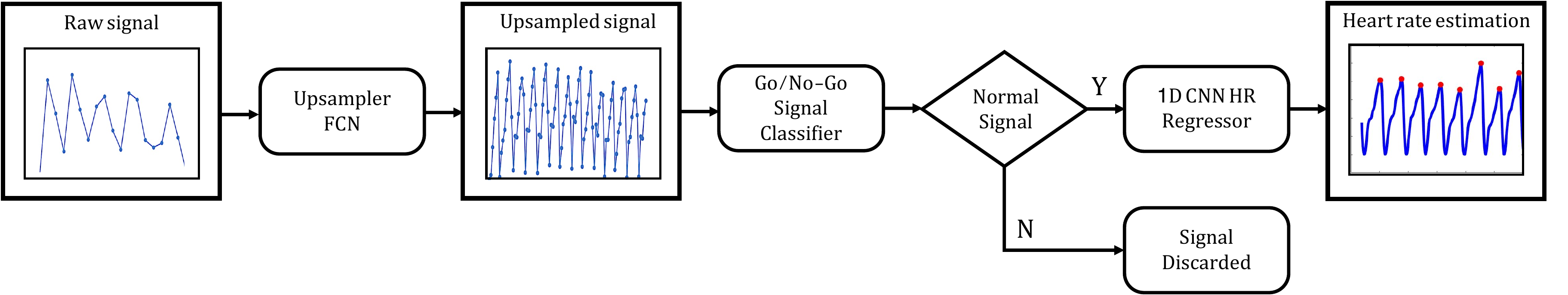}
\caption{Complete Machine learning pipeline consisting of an upsampler neural network, a classifier to prevent abnormal signals from further processing and a heart rate regression neural network designed to directly predict heart rate from the dataframe.}
\label{Fig_2: ml_pipeline}
\end{figure*}

The main constituents of the proposed ML pipeline and the flow of data are shown in Fig. \ref{Fig_2: ml_pipeline}. The pipeline consists of three parts, upsampler, classifier and HR regressor. The architecture and parameters associated with each of the constituents are discussed below. The pipeline is assembled along with their tensors and matrices after the individual models were trained and validated in the format discussed below.

\subsection{Dataset for ML pipeline}
The dataset is split into 3 parts: Training data (70\%), Validation data (15\%) and Test Data (15\%). These three splits are kept the same while training the upsampler, classifier and regressor. The regressor is only trained with the upsampled normal signals since the classifier would reject abnormal signals when the system is deployed and hence they would not be passed as input to the regressor.

\subsection{Upsampler}

The objective of the Upsampler is to reconstruct the signal to its original version for a given low-dimensional input. We employ a Fully Connected Neural Network (FCN) to achieve this.

The architecture is composed of three layers; an Input layer with 35 neurons, a hidden layer and an output layer with 69 neurons (Table \ref{table:upsampler_arch}). 'Relu' (Rectified Linear Unit) activation function was used in the neural network. The model performance was evaluated using Root Mean Squared Error (RMSE). The model is trained using stochastic gradient descent with a learning rate of 0.9 for 2000 epochs. Different runs are performed by varying number of neurons in the hidden layer from 30 to 55. The network with 55 neurons in the hidden layer achieved the best test RMSE (0.094). 
The parameters of the FCN are updated using stochastic gradient descent and the Mean Squared Error (MSE) loss function was used to compute the gradient.

\subsection {Classifier}
The classifier is trained to predict if a signal is normal or abnormal. 
A 1-D CNN architecture with two hidden layers was used for the classifier. Table \ref{table:classifier_regressor_arch} shows the architecture of the classifier. The abnormal and normal signals were labelled positive and negative respectively.  The network is trained using the binary cross entropy loss function. For gradient descent, the Adam \cite{kingma2014method} optimization algorithm is used with a learning rate of 0.003. The model is trained for 1000 epochs. The model checkpoint with the smallest F1-score on the validation set was chosen to report the final results. The performance of the classifier is evaluated using its accuracy and F1-score on the test set. F1-score is the harmonic mean of precision and recall metrics.

\subsection {HR Regressor}
We perform our experiments using the FCN and 1-D CNN architecture.  The input layer has 69 neurons and the output layer has one neuron to yield the heart rate value. The output from the best performing upsampler is provided as input to the HR Regressor. Both the hidden layers in the FCN have 8 neurons each with the sine activation function between these layers. Table \ref{table:classifier_regressor_arch} depicts the architecture of the 1-D CNN. For both types of networks, we use Adam as the Gradient Descent optimization algorithm with Mean Squared Error (MSE) as the loss function. A learning rate of 0.05 and weight decay of 0.05 is used, and the model is trained for 1000 epochs. The model checkpoint with the smallest RMSE on the validation set was chosen to report the final results. The evaluation metrics used for the regressor are RMSE and MAE.

\subsection{Details of Experiments}
Hyperparameter tuning was performed to determine the learning rate, weight decay, the number of hidden neurons for each layer (FCN), output channels (1-D CNN) and kernel size for the convolution layers (1-D CNN). For all three components of the ML pipeline, shallow architectures were chosen such that their combined size is similar to the size of the signal processing pipeline on the edge device. 
In Table \ref{table:test_results}, the standard deviation for the evaluation metrics is reported by running each experiment with 5 different random seeds that result in different weight initialization for the models. 

\subsection{Hybrid Pipeline}
The proposed hybrid pipeline that combines the U-NN, signal quality classifier from the ML pipeline with the peak detection and HR estimation algorithms from the signal processing pipeline were developed and tested to take advantage of the strengths of both approaches. Fig. 3. shows the flow of data and the constituents of the hybrid pipeline. The hybrid pipeline enables us to take advantage of the efficiency of the ML pipeline while also benefiting from the high inference accuracy of the traditional peak detection algorithm for HR estimation. 

\subsection{Deployment on Edge Device}

The three different pipelines proposed in the current work were deployed on an ESP32 edge device. The inference time, energy consumption and inference accuracy were measured on edge. The deployment on ESP32 was facilitated by converting PyTorch models of the U-NN, Classifier and HR-NN into ONNX. The ONNX models were then ported over to TensorFlow and to TFLite, to make the models compatible with ESP32. The porting over process was carried out to remove all overheads, unnecessary libraries and dependencies that could reduce the inference efficiency on edge.

\begin{figure*} [!h]
\centering
\includegraphics[width=\linewidth]{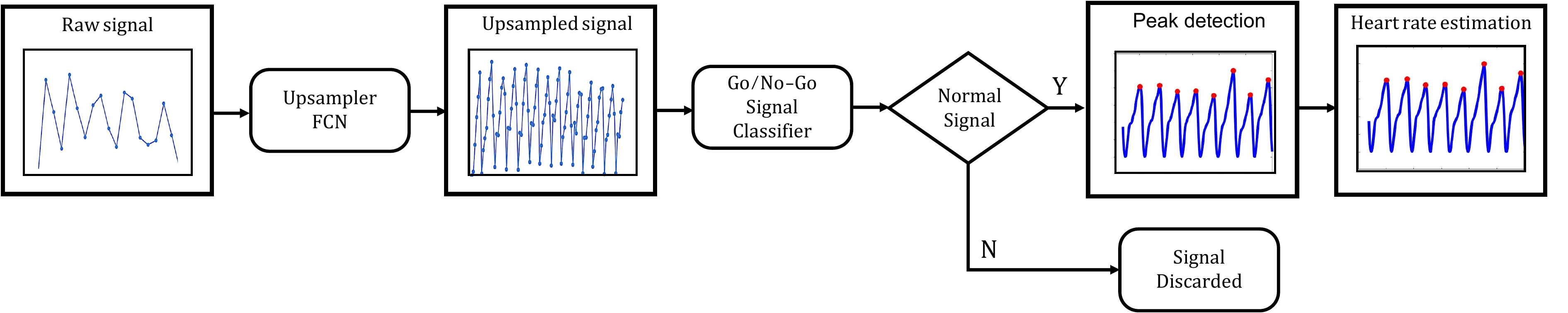}
\caption{Hybrid signal processing pipeline consisting of an upsampler neural network, a classifier to prevent abnormal signal from further processing and a traditional peak detection algorithm designed to count the total number of peaks within a dataframe.}
\label{Fig_3: hybrid_pipeline}
\end{figure*}

\section{Results and Discussions}

\subsection{Signal Requirements for Heart Rate Estimation}

Heart rate estimation is a relatively simple process that involves peak detection and peak-to-peak distance estimation to determine the mean value over a sliding time window. This process leads to accurate heart rate estimations when a good quality input signal with well-defined peak morphology is used. The Nyquist frequency of the peak features used to estimate heart rate ranges from 0.3 to 3 Hz for healthy subjects, which provides a theoretical guarantee that the feature can be reconstructed at sampling rates above 6 Hz. In practice, the sampling rate as high as eight times Nyquist frequency is used to ensure sufficient signal quality and accuracy. However, this approach leads to long inference times, higher energy consumption and computational requirements on edge devices. For optimal accuracy and energy usage characteristics on edge, a sampling rate of 12 Hz can be used, with a suitable signal processing pipeline that can reliably predict heart rate under most conditions. Such a signal processing pipeline would have to incorporate signal conditioning, quality classification, non-linear interpolation and feature extraction. These steps increase the inference time and energy consumed by the edge device. In this study, we demonstrate a proof of concept machine learning (ML) pipeline that overcomes the limitations of traditional signal processing by reducing the inference time and energy demand. Further, we compared the accuracy characteristics of the proposed ML model  versus a lightweight signal processing algorithm. 

Fig. \ref{Fig_3: hybrid_pipeline} shows the flow of data through the complete ML pipeline, consisting of the U-NN, signal quality classifier and the HR-NN that infers the heart rate. These models have the advantage of being more interpretable than end-to-end machine learning models. In comparison to our proposed ML pipeline, the traditional signal processing algorithm had a slightly higher accuracy, but it required a much higher inference time leading to higher energy usage.

\begin{table}[]
\centering
\renewcommand{\arraystretch}{1.3}
\begin{tabular}{|c|c|c|c|}
\hline
\textbf{\begin{tabular}[c]{@{}c@{}}Input  Size\end{tabular}} & \textbf{Layer} & \textbf{\begin{tabular}[c]{@{}c@{}}Number of \\ neurons\end{tabular}} & \textbf{\begin{tabular}[c]{@{}c@{}}Activation \end{tabular}} \\ \hline
35                                                             & Linear         & 55                                                                    & ReLU                                                                   \\ \hline
55                                                             & Linear         & 69                                                                    & -                                                                      \\ \hline
\end{tabular}
\caption{2-layer Fully Connected network architecture for Upsampler}
\label{table:upsampler_arch}
\end{table}

\begin{table}[]
\renewcommand{\arraystretch}{1.5}
\begin{tabular}{|c|c|c|c|c|}
\hline
\textbf{Input Size} & \textbf{Layer}    & \textbf{Kernel Size} & \textbf{Output Size} & \textbf{Activation} \\ \hline
69x1       & 1-D Conv & 5           & 65x5        & ReLU       \\ \hline
65x5       & 1-D Conv & 5           & 61x5        & ReLU       \\ \hline
305        & Linear   & -           & 1           & Sigmoid    \\ \hline
\end{tabular}
\caption{Same 3-layer 1-D CNN architecture is used for the Classifier and the HR Regressor. The Regressor has the Sine activation function instead of ReLU and it does not have a sigmoid layer at the end.}
\label{table:classifier_regressor_arch}
\end{table}

\subsection{Signal Conditioning and Upsampling}

\begin{figure}[!h]
\centering
\includegraphics[width=\linewidth]{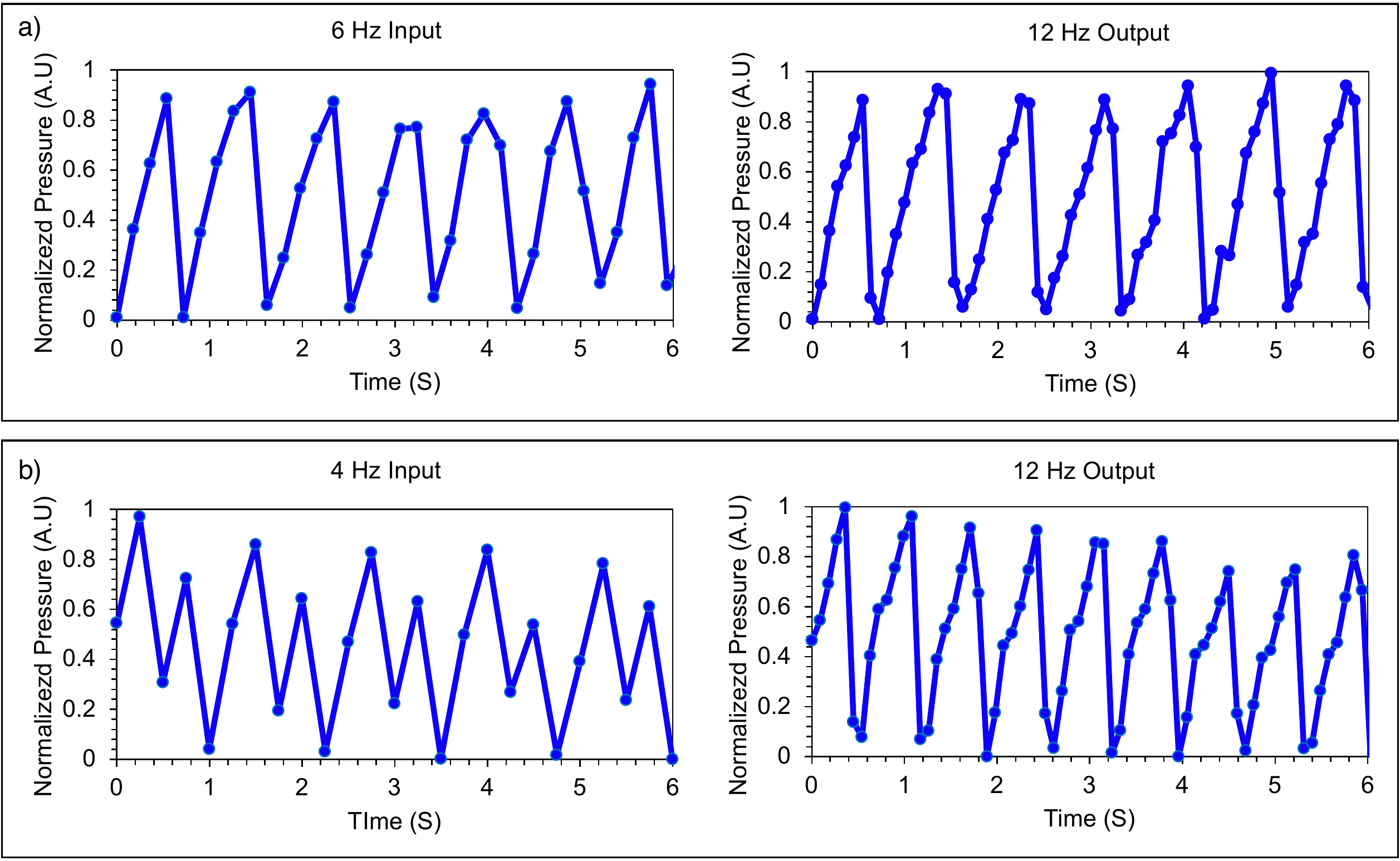}
\caption{Plots showing the reconstructed pulse signals obtained from upsampling-NN with a) 6 Hz and b) 4 Hz raw samples as inputs and the resultant 12 Hz output signal. The U-NN was trained on 12 Hz ground truth signals that were processed using signal conditioning and smoothing algorithms. The image clearly show that the U-NN, in addition to upsampling, reliably reconstructs the features of the raw signal.  }
\label{fig_5: reconstructed_signals}
\end{figure}

Signals, when acquired at low sampling rates, require substantial conditioning and upsampling to extract meaningful features using traditional signal processing. This results in higher computational needs and longer processing times which makes low sampling rates unattractive for most applications. Further, the use of simple interpolation and curve fitting leads to signal distortions compromising the prediction accuracy. The U-NN proposed in the current study combines the functions of signal conditioning and interpolation, effectively reconstructing a realistic representation of the ground truth signal acquired at higher sampling rates. Fig. \ref{fig_5: reconstructed_signals} shows representative signals reconstructed from 4 and 6 Hz to 12 Hz by the U-NN. The RMSE between the reconstructed 12Hz signal and the ground truth 12 Hz signal was 0.094, which indicates good reconstruction quality and agreement with the ground truth. Considering the constraints of edge deployment, we chose a shallow FCN architecture with a model size of only 22.8 kB and consists of 5844 parameters which are well within the flash memory and RAM requirements of ESP32. Table 1 shows the relevant metrics of the U-NN.

\subsection{Signal Quality Classification and Quality Heuristics}

\begin{figure}[!h]
\centering
\includegraphics[width=\linewidth]{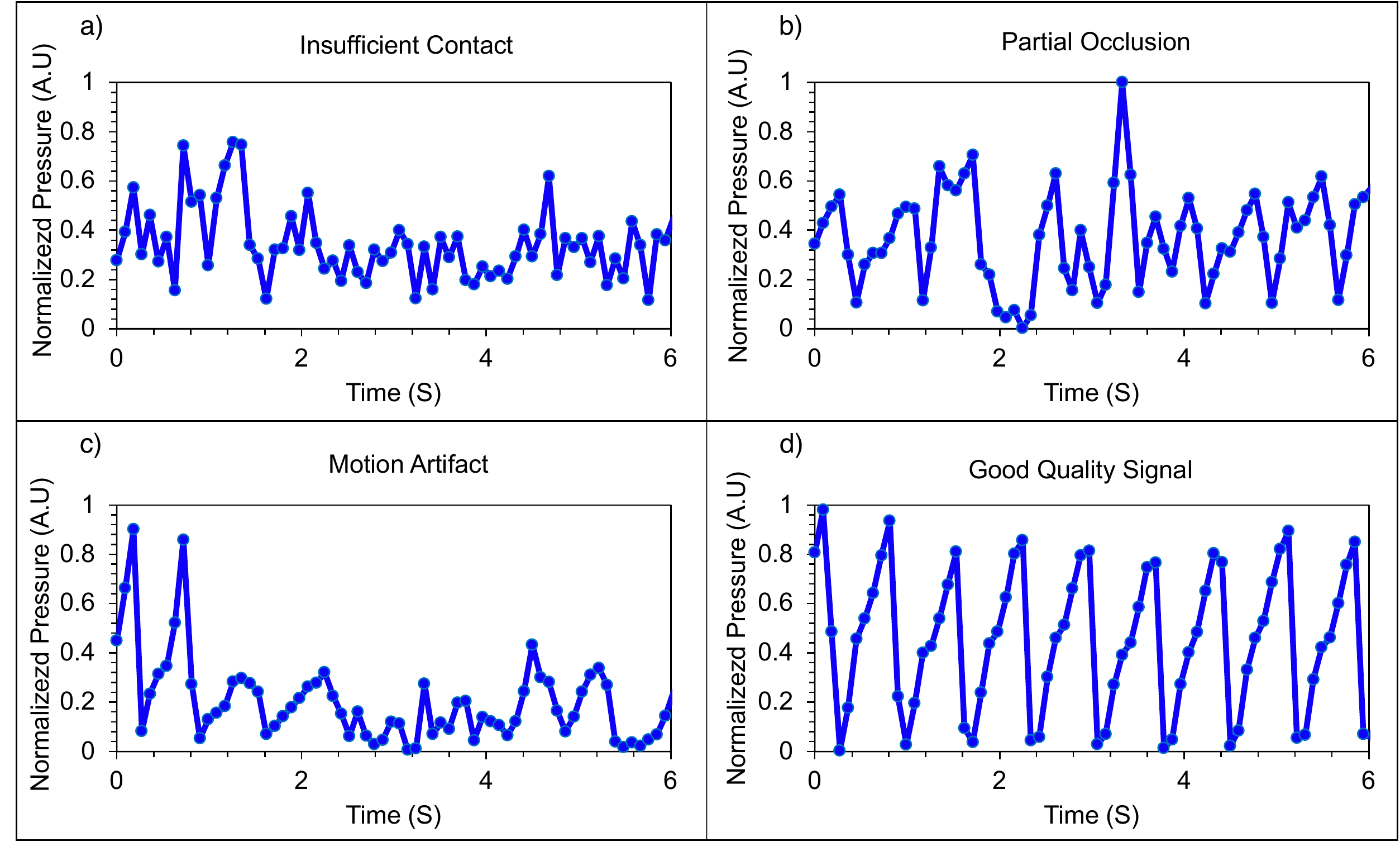}
\caption{Typical signals that are classified a good versus bad based on the signal quality and peak prominence criteria. The good signals are passed to the regression step where the heart rates are estimated. }
\label{fig_6: good_poor_class}
\end{figure}

The signal quality is one of the most important determinants of heart rate estimation accuracy. Factors such as insufficient sensor contact, motion artifacts, and partial occlusion of the blood vessels can affect the signal and make them unusable for carrying out inferences. Many studies in the literature have used classification models to improve signal processing efficiency and to avoid processing datasets that are not suitable. Using traditional signal processing techniques, such signal quality classification can be achieved by using heuristics and rule-based decision-making algorithms that can identify abnormal data and eliminate them. In addition, post inference rules which consist of physiological range bounds, and outliers can be used for anomaly detection. 
In our study, we have developed a 1D CNN model that classifies data with a binary label based on the signal quality. The classifier acts as a go/no-go gate that prevents poor quality signals from being used for inferences. Fig \ref{fig_6: good_poor_class} shows some common examples of datasets classified as having poor quality. The accuracy and F1-Score of the classifier were found to be 94\% and 0.72 respectively when exposed to the test dataset. 1D CNN being a compact architecture resulted in a model size of only 1.82 kB and 466 parameters. Table \ref{table:test_results} shows the relevant metrics of the classifier.

\begin{table*}[]
\centering
\renewcommand{\arraystretch}{1.5} 
\begin{tabular}{|c|c|c|c|cc|}
\hline
\textbf{Model} & \textbf{Architecture} & \textbf{\begin{tabular}[c]{@{}c@{}} Model Size\\ (kB) \end{tabular}} & \textbf{\begin{tabular}[c]{@{}c@{}}Number of \\ Parameters\end{tabular}} & \multicolumn{2}{c|}{\textbf{Performance Metrics}}                    \\ \hline
Upsampler      & FCN  (Table \ref{table:upsampler_arch})                 & 22.8                                                               & 5844                                                                     & \multicolumn{2}{c|}{RMSE = 0.094$\pm$0.002}   \\ \hline
Classifier     & 1-D CNN (Table \ref{table:classifier_regressor_arch})              & 1.82                                                               & 466                                                                      & \multicolumn{1}{c|}{Accuracy = 94$\pm$0.3\%} & F1-Score = 0.72$\pm$0.008 \\ \hline
Regressor      & 1-D CNN (Table \ref{table:classifier_regressor_arch})             & 1.82                                                               & 466                                                                      & \multicolumn{1}{c|}{RMSE = 4.8$\pm$0.07}     & MAE = 3.28$\pm$0.14       \\ \hline
\end{tabular}
\caption{Test Set Results of different components of the ML pipeline.}
\label{table:test_results}
\end{table*}

\subsection{Heart Rate Estimation}

The final set of steps in the pipeline to extract heart rate estimations involve the pulse peak detection and peak-to-peak distance measurement for the hybrid and the traditional processing pipeline. The accuracy of heart rate estimates is directly affected by these features. In a signal processing pipeline, the feature extraction consists of filters and algorithm parameters that have to be hand-tuned and optimized to ensure accuracy. These parameters for commonly extracted features are usually set based on predetermined rules and heuristics. The lightweight signal processing pipeline that we created using these rules performed well in terms of accuracy with an MAE of 1.13 and RMSE of 3.44. The size of the entire signal processing pipeline was 22 kB and leaving open the possibility of further improving the accuracy of heart rate predictions using more sophisticated approaches. Table \ref{table:esp_comparison} shows relevant metrics for the signal processing pipeline. 

\begin{figure*} [!h]
\centering
\includegraphics[width=\linewidth]{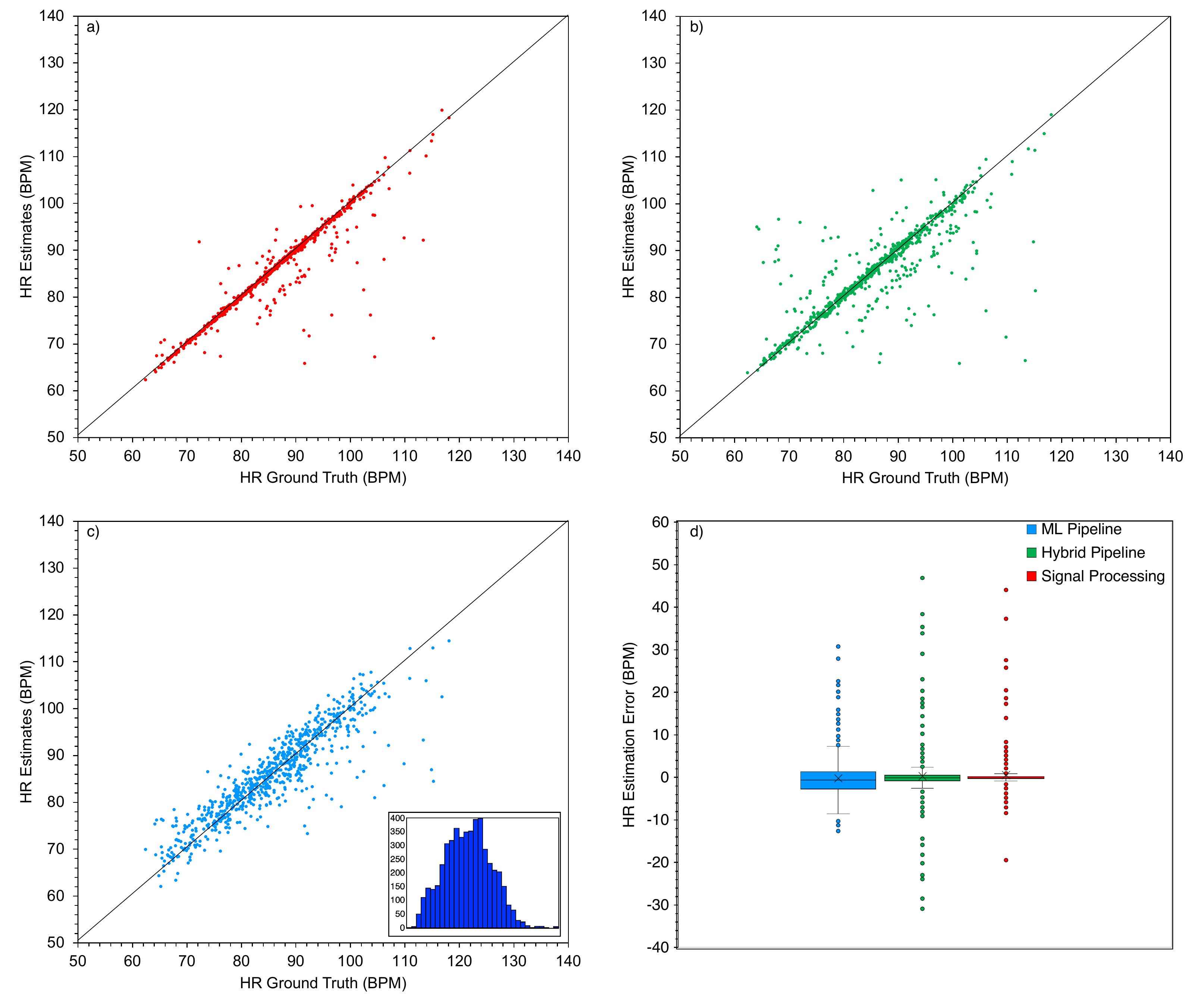}
\caption{Plots showing the statistical inference error of heart rate estimates made by the ML, Hybrid and signal processing pipelines, compared with ground truth HR obtained by measuring peak to peak time. The plots show only the test set data. a) The HR estimates made by signal processing which shows good agreement with the ground truth. b) Hybrid pipeline HR estimates Vs. ground truth, which shows a slight degradation in estimation accuracy. c) ML pipeline HR estimates Vs. ground truth, showing good fit, albeit with a larger mean error. d) Shows the box and whisker plot with outliers, indicating that the interquartile range for signal process is the narrowest with hybrid and ML pipelines having slightly larger deviation. However, the ML pipeline does not make as many outlier estimates as signal processing and hybrid HR predictors. The black diagonal lines represent the ground truth HR values.}
\label{fig_7: hr_pred_comparison}
\end{figure*}

In our proposed ML pipeline, the 1D CNN heart rate regressor uses the reconstructed 12 Hz signal as input and provides a signal HR value as output. Tables \ref{table:classifier_regressor_arch} and \ref{table:test_results} shows the architecture and relevant statistics of the regressor respectively. The model size is only 1.8 kB and has 499 parameters. The 1D CNN through learning, abstracts the relevant features necessary for peak detection and HR estimation, as compared to the traditional peak detection algorithms. Thus, the dependence on hand tuning of parameters is eliminated in the ML pipeline. Table \ref{table:esp_comparison} shows the performance and accuracy metrics of the ML, hybrid and the signal processing pipeline, compared in this study. The results show that the ML pipeline has a much lower inference time (nearly 6x) and consequently, lower energy consumption. However, the inference accuracy (MAE = 3.28 BPM) was found to be lower than both hybrid (MAE = 2.1) and signal processing (MAE = 1.17) pipelines. Further, the HR estimation error was found to be similar for both the train and test set, which suggests that the CNN has not over-fitted to the training dataset. The visual representation of the HR predictions made by the three models versus the HR ground truth is shown in Fig. \ref{fig_7: hr_pred_comparison}. The fig. \ref{fig_7: hr_pred_comparison}.a), b) and c) shows the plot of HR estimation Vs. HR ground truth. Further, the ML pipeline results in estimations that have a wider inter-quartile range and standard deviation than both hybrid and signal processing as shown in Fig. \ref{fig_7: hr_pred_comparison}.d). However, the total number of outlier HR estimates is much lower for the ML pipeline. This result is in agreement with prior studies \cite{etiwy2019accuracy} that have shown that signal processing algorithms deployed in commercial edge devices tend to make large estimation errors from time to time, due to misidentification of peaks or due to poor signal quality. The ML pipeline, on the other hand, does not make large estimation errors as long as the real-world data falls within the distribution of the data used to train the models. The potential and limitations of the ML pipeline are elucidated further in the discussions section.

\begin{table*}[]
\centering
\renewcommand{\arraystretch}{1.5} 
\begin{tabular}{|c|c|c|c|}
\hline
                         & \textbf{Signal Processing Pipeline} & \textbf{Hybrid Pipeline} & \textbf{ML Pipeline} \\ \hline
Model size               & 22 kB                               & 35 kB                    & 39 kB                \\ \hline
Inference time           & 2.51 ms                              & 1.82 ms                   & 0.41 ms              \\ \hline
Estimated inference energy         & 7.4 mJ                             & 5.15 mJ                   & 1.21 mJ              \\ \hline
Mean absolute error of entire data set (BPM) & 1.17                             & 2.35                  & 2.99              \\ \hline
Mean absolute error of test set (BPM)      & 1.17                                & 2.39                     & 3.28                  \\ \hline
\end{tabular}
\caption{Comparison of the model size, inference time, inference energy consumption, ADC current draw and inference accuracy on ESP32 edge device for the traditional signal processing, hybrid and complete ML pipelines. The model size includes the libraries used to run the models, in addition to the models themselves. All inferences were conducted at a clock speed of 40 MHz.}
\label{table:esp_comparison}
\end{table*}

\section{Discussions}

\subsection{Generalizability}

One of the major challenges for ML models is the lack of generalizability when exposed to data that is qualitatively different from the training set. All the models proposed in this study have been trained on datasets considered to be in the normal range. Although, all the models both individually and when combined in an edge deployed pipeline, perform well in terms of prediction accuracy and low error, could fail when they encounter signals with characteristics different from the training datasets. The go/no-go classifier was designed to, at least partially address this challenge, wherein the classifier will identify abnormalities in the sample and prevent them from being used for inferences and HR predictions. However, even with 94\% accuracy, the classifier allows some abnormal signals to pass through to the inference stage. The generalizability of ML models is an active area of research and is a known impediment to the deployment of ML models in real-world applications. One potential solution is to develop more advanced models and train them on diverse datasets with more abnormal signals to improve F1-Score and accuracy. Currently, the only way to ensure improved generalizability is to train the models on a much wider distribution of data, which will be the focus of our future studies.

\subsection{Accuracy}
The accuracy of the proposed pipeline, while promising, is not theoretically guaranteed as is the case with traditional signal processing. For greater generalizability, and accuracy in real-world conditions, much larger training datasets have to be acquired to train and validate the ML models, which might be impractical and expensive to achieve. This is one of the reasons we developed the hybrid pipeline, which shares most of the ML pipeline, while deviating only at the HR inference stage, where traditional signal processing is used, instead of the HR regressor. Due to the relatively small model size of the regressor, we can deploy both the HR regressor and the signal processing HR estimator to enable potential ground truth acquisition on the edge device, without compromising much on the energy and inference characteristics. This has the added benefit of introducing naturally occurring variability in the datasets, improving the inference accuracy through federated learning over time.

\subsection{Interpretability of the pipeline}
One of the biggest problems with end-to-end machine learning is the lack of interpretability and the general black box nature of the models. The proposed pipeline partly addresses this by introducing probing points within the pipeline that enables an expert to view the outputs of the models to determine how they are performing from an application function standpoint. The FCN and 1D CNNs themselves are not interpretable, however, the results from these networks can be analyzed and probed to determine how well they are performing in real-world scenarios. The output of the upsampler can be visually and mathematically verified to determine how well it is reconstructing the suboptimally sampled signals. The classifier also acts as a quality check for the reconstructed signal. If the U-NN performs a poor reconstruction for an abnormal or low-quality signal, the classifier is trained to label it as a poor-quality signal and prevent it from further processing. Therefore, we have introduced multiple steps designed to prevent an HR inference which is wrong by a wide margin. The ML pipeline, thus, by design is biased towards not reporting an HR value when it encounters a signal which might lead to a prediction with a large error.

\subsection{Limitations}
Despite the promising results of the ML pipeline, there are limitations that have to be addressed before considering their deployment on edge devices. The current study is strictly a proof of concept designed to demonstrate the inference time and energy savings possible by using cutting-edge lightweight neural networks while making them more interpretable than black box models. Therefore, we have made some compromises in the present study, namely, only 5 volunteers were used for the preliminary training and test datasets and a comprehensive network architecture search has not been used to determine the most optimal ML models. While we have used the model that balances the accuracy and efficiency requirements of the edge devices, a more optimal model might be found if a deeper architecture search is carried out. We plan to address these challenges in our future studies using more volunteers and carrying out a network architecture search to further optimize the models.

\section{Conclusion}
We have demonstrated that a three-part ML pipeline consisting of a shallow FCN and 1D CNNs can be used to estimate HR values from a pressure pulse signal. The pipelines were deployed in an ESP32 edge device and the performance characteristics were determined. When compared to traditional signal processing, the ML pipeline was nearly 6x more efficient in inference times, while being only slightly inferior in terms of inference accuracy. We further show that the ML pipeline is more interpretable than an end-to-end ML model by breaking out the functional aspects of the HR estimation problem into upsampling, classification and HR regression, which enables us to probe the inferences made by the models. Further, we constructed a hybrid pipeline, combining the efficiency advantages of the ML pipeline with the interpretability and higher accuracy of the HR estimation algorithms. All three models discussed in this study can be deployed in resource-constrained environments with each affording complementary advantages and tradeoffs. Being modular, the elements of the models can be mixed and matched based on system and application constraints.

\blfootnote{Code link - \href{https://github.com/tataganesh/HRV-edgedevice.git}{\textcolor{blue}{https://github.com/tataganesh/HRV-edgedevice.git}}} 

\bibliographystyle{IEEEtran}
\bibliography{References.bib}

\newpage

\vfill
\end{document}